%% file: STSG_manuscript_arXiv.tex
\documentclass[12pt]{amsart}
\usepackage{amscd,amssymb,mathrsfs,amsmath,hyperref,algorithmicx,algpseudocode,algorithm,eqparbox,array}%cite\usepackage{hyperref} algorithm2e
\usepackage{multirow,graphicx,epstopdf,bbm}
\usepackage{xcolor}
\usepackage{endnotes}

\usepackage[margin=2cm]{geometry}
\usepackage{setspace}

\usepackage{color}
\usepackage{xparse}

\input{Defs.tex}

\begin{document}

\title[]
{Machine olfaction using time scattering of sensor multiresolution graphs}

%%    Information for first author
\author[]{Leonid Gugel}
\address{School of Mathematical Sciences,
Tel Aviv University,Ramat Aviv,
 6997801 Tel Aviv, Israel}
 \email{leong@post.tau.ac.il}
%\urladdr{http://www.math.biu.ac.il/$^\sim$gugoll}

%%    Information for first author
\author[]{Yoel Shkolnisky}
\address{School of Mathematical Sciences,
Tel Aviv University,Ramat Aviv,
 6997801 Tel Aviv, Israel}
 \email{yoelsh@post.tau.ac.il}
\urladdr{https://sites.google.com/site/yoelshkolnisky/}

%%    Information for second author
\author[]{Shai Dekel}
\address{School of Mathematical Sciences,
Tel Aviv University,Ramat Aviv,
 6997801 Tel Aviv, Israel}
\email{shai.dekel@ge.com}
\urladdr{http://www.shaidekel.com/}

\date{February 11, 2016}

\include{Abstract}

\keywords{sensor array, electronic nose, metal-oxide sensors, odor classification, odor location, classification of high dimensional time series, scattering, deep learning, random forest}

%\thanks{This work is }

\maketitle

\tableofcontents

\include{Introduction}

\include{MainPart}

%%%%%%%%%%%%%%%%%%%%%%%%%%%%%%%%%%%%%%%%%%%%%%%%%%%%%%%%%%%%
% \vskip 0.7cm
%\bibliography{leon_bib}{}
\bibliography{STSG_manuscript_arXiv.bbl}{}
\bibliographystyle{amsplain}
%\begin{thebibliography}{10}
%\end{thebibliography}
%\print

%bibliography

\end{document}

%% file: Defs.tex
\newcommand{\rvc}[2]{

\textcolor{red}{#1}%
    \IfNoValueF{#2}{\protect\endnote{#2}%
    }%
}

 \numberwithin{equation}{subsection}

\newcommand\LONGCOMMENT[1]{%
  \hfill\#\ \begin{minipage}[t]{\eqboxwidth{COMMENT}}#1\strut\end{minipage}%
}

\newtheorem{thm}{Theorem}[section]

 \theoremstyle{definition}

\newtheorem{note}[thm]{Notes}
\theoremstyle{remark}

\numberwithin{equation}{section} %%

\def\E {{\mathbb{E}}}
\def\R {{\mathbb{R}}}

\def\Z {{\mathbb{Z}}}
\def\T {{\mathbb{T}}}

\def\v {\mathcal{V}}
\def\vv {\mathscr{V}}
\def\f {\mathcal{F}}

\def\Ind {\mathbbm{1}}

\def\l {\ell}

\newcommand{\ben}{\begin{enumerate}}

\newcommand{\een}{\end{enumerate}}
\newcommand{\bit}{\begin{itemize}}
\newcommand{\eit}{\end{itemize}}

\def\x{\times}

\def\+{\oplus}
\def\!={\neq}
\def\=<{\leqslant}
\def\>={\geqslant}
\def\...{\ldots}
\def\<{\prec}

\def\QED{\nobreak\quad\ifmmode\roman{Q.E.D.}\else{\rm Q.E.D.}\fi}

\def \Exp#1{e^{#1}} %\mathrm{e}

\def\~dl{\overline{ \delta }}

\def\-->>{\Rightarrow}
\def\<<--{\Leftarrow}

%% file: Abstract.tex
\begin{abstract}
In this paper we construct a learning architecture for high dimensional time series sampled by sensor arrangements. Using a redundant wavelet decomposition on a graph constructed over the sensor locations, our algorithm is able to construct discriminative features that exploit the mutual information between the sensors. The algorithm then applies scattering networks to the time series graphs to create the feature space. We demonstrate our method on a machine olfaction problem, where one needs to classify the gas type and the location where it originates from data sampled by an array of sensors. Our experimental results clearly demonstrate that our method outperforms classical machine learning techniques used in previous studies.
\end{abstract}

%% file: Introduction.tex
\section{Introduction}

Developing chemo-sensing solutions and standards for early warning against chemical and biological hazards   has been an active research area~\cite{chemHazards,guideHazards}. To construct an accurate and reliable chemical warning system, the information from several sensors must be integrated to provide a clear indication for the composition of the chemical substances as well as their propagation profile. We propose a machine learning approach, that is based on recent advances, for classification and regression problems in machine olfaction. Machine olfaction problems include odor classification problems, gas consecration detection, and chemical source localization.

Our algorithm consists two steps of feature generation and classification. To transform the raw data of the sensor array platform into discriminative features, we propose the Scattering Time Series on Graphs(STSG) transform, which is an hierarchical feature extraction method from multiple time series. This transform is an extension of the recently proposed scattering transform for time series~\cite{mallat2012group,bruna2013invariant,bruna4104multiscale,bruna2013audio} and graph signals~\cite{chen2014unsupervisedHaarScatGraphs}, to multivariate time series defined on a graph. The resulting features are then classified using a random forest (RF) based classifier~\cite{breiman2001random, breiman1984classification}.

We demonstrate and evaluate our algorithm by means of the “Dataset from chemical gas sensor array in turbulent wind tunnel”~\cite{fonollosa2015dataset}, available through the UCI Machine Learning Repository~\cite{webUCIgasSensors}.
This dataset has been used to validate existing algorithms~\cite{vergara2013performance, vembu2012time}.

The structure of our paper is as follows. In Section~\ref{ch:dataset} we describe a set of machine olfaction problems we wish to solve and the dataset that are used. In Section~\ref{sec:sateOFart} we review prior art methods.

In Section~\ref{ch:Math}, we present the theoretical building blocks of our method: a redundant Haar wavelet decomposition over a (possibly irregular) graph, the scattering convolution network and Random Forests. Equipped with these building blocks, we present in Section~\ref{ch:MainAlgo} the main contribution of this paper, the \emph{\textbf{STSG}} (Scattering of a Time Series on a Graphs) algorithm.

Finally, in Section \ref{ch:Exper}, we show experimental results of our methodology applied in the machine olfaction setting: odor discrimination, odor consecration, and odor localization. We compare the performance of our method with prior techniques of machine olfaction~\cite{vergara2013performance,vembu2012time}.

%% file: MainPart.tex
\section{Olfaction datasets} \label{ch:dataset}
In this paper we use the dataset ``Gas sensor arrays in open sampling settings" from the UCI archive (Machine learning Repository)~\cite{webUCIgasSensors}. This dataset is a collection of multidimensional time series data along with some static environmental parameters that include the responses of a chemical detection platform to different gases at different levels of concentration. The challenge is to develop machine learning techniques for gas classification and source prediction models. In this section, we review the dataset in details.

\begin{figure}
  \centering
  \includegraphics[scale=0.5]{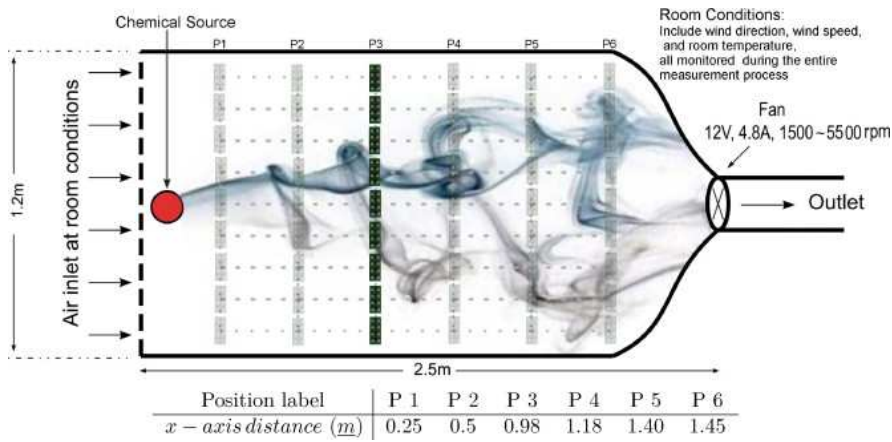}\\
  \caption{Wind tunnel used to collect time series data from sensor arrays\cite{fonollosa2015dataset}.}\label{fg:Wind_tunnel}
\end{figure}

The data in~\cite{webUCIgasSensors} was collected in a  $2.5m \times 1.2m \times 0.4m$ wind tunnel test-bed facility (see Figure~\ref{fg:Wind_tunnel}), into which the gaseous substances of interest were released.

The wind tunnel operates in a propulsion open-cycle mode, by continuously drawing external turbulent air throughout the tunnel and exhausting it back to the outside, creating a relatively less turbulent airflow moving downstream towards the end of the test field. In order to construct various distinct artificial airflows in the wind tunnel, the wind tunnel contains a motor-driven exhaust fan at the outlet of the test section. The motor can be set to rotate at three different constant speeds. The wind tunnel measures the ambient temperature and relative humidity during the entire experiment.

\begin{figure}
  \centering
  \includegraphics[scale=0.75]{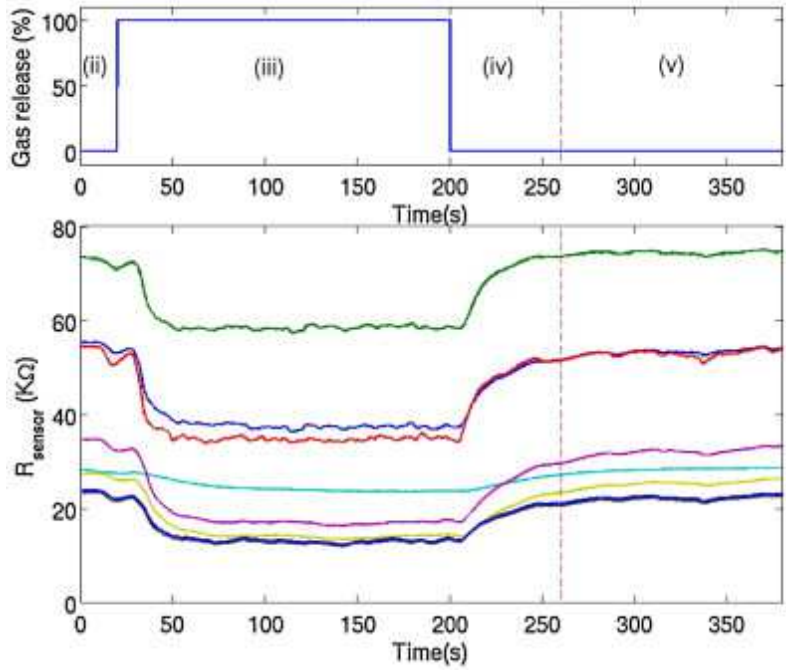}\\
  \caption{Multivariate response of a 8-sensor array when methane is released in the wind tunnel~\cite{fonollosa2015dataset}.}\label{fg:response}
\end{figure}

The chemical detection platform inside the wind tunnel consists of boards. Each board has eight commercialized metal-oxide gas sensors (MOX)~\cite{FigaroIncSite}, which are sensitive to rapid changes in the analytes concentration. Thus, the output of each board is an $8$-dimensional time series. The chemical detection platform consists of columns of nine boards each located at six equally-spaced positions along the wind tunnel, that is, a total of $72$ sensors per location (see Figure~\ref{fg:Wind_tunnel}). Figure~\ref{fg:response} depicts a typical time-series response of one board. The sensor responses are affected by the air turbulence in the wind tunnel and depend on the concentration of the gas substance. As the operating temperature of the sensors affects their performance, it is adjustable by setting the voltage of the built-in heater of each sensor to one of five different levels.

The dataset~\cite{webUCIgasSensors} was generated by releasing ten different types of gas into the wide tunnel: acetone, acetaldehyde, ammonia, butanol (butyl-alcohol), ethylene, methane, methanol, carbon monoxide, benzene, toluene, and carbon monoxide(CO).

Each kind of chemical substance is released at same  nominal  concentration  values  at  the  outlet  of  the gas  source  in  parts-per-million  by  volume(ppmv).The CO  was released in two different nominal concentrations. The value of  concentration at gas  source is from $100$ ppm to $400$ ppm. Note that the actual concentration in the wind tunnel decreases as the generated gas plume spreads out along the tunnel.

For each gas released, the motor of the exhaust fan was set to one of three rotation speeds. A  turbulent airflow was thereby generated within the wind tunnel.
The outputs of the sensors at each of the six locations in the wind tunnel were measured separately, resulting in six $72$-dimensional time series datasets capturing the chemical analyte circulated throughout the wind tunnel. The temperature of all sensors remained fixed during the measurement. After each test, the wind tunnel was thoroughly ventilated. The measurement was repeated for three different air velocities and five operating temperatures.
 For each combination of the type of gas, airflow velocity, and temperature, the measurement (generation of time-series) was repeated $20$ times. The time-series measure at each location was sampled for approximately $250$ seconds, with $10$ samples per second, that is, a total of about $2500$ samples per location per experiment. The time series measured at all locations was fully synchronized, although each was of a slightly different length. The data for each of the ten gas classes was collected $20$ times for each three speeds, five temperatures modes and six locations. The resulting dataset thus consists of 18,000 72-dimensional time series.

   In addition, each experiment recorded the ambient temperature and relative humidity during the experiment. Although these parameters are also represented as time series, they have very low variance and can thus be represented by their average values. For a more detailed description of the experimental protocol see~\cite{fonollosa2015dataset,vergara2013performance}.

%\begin{table}[h]
%\caption{Sensor position distance from chemical source}
%\centering % used for centering table
%\begin{tabular}{c|c}
%\hline\hline %inserts double horizontal lines
%Position label & Axis  distance \\[0.5ex] % inserts table
%\hline\hline %inserts single line
%P1             & 0.25           \\\hline
%P2             & 0.5            \\\hline
%P3             & 0.98           \\\hline
%P4             & 1.18           \\\hline
%P5             & 1.40           \\\hline
%P6             & 1.45\\\hline\hline% [1ex] adds vertical space
%
%\end{tabular}
%\label{tbl:Pos_loc}
%\end{table}
%By obtaining of we data we know location on each time series: position and board number.
%There exist six training position \wrt chemical source are labeled in Figure \ref{fg:Wind_tunnel}.
% are shown in Table \ref{tbl:Pos_loc}.
%The sensing board are located by the center of 9 board position which is uniform array $0.13:0.13:1.2$

\section{Related studies on odor classification}\label{sec:sateOFart}
Several previous studies~\cite{nimsuk2007improvement, nimsuk2008study, vembu2012time, vergara2013performance} have considered the problem of odor classification. These studies all follow the same approach of extracting features from the data followed by some classification scheme.

Earlier approaches of odor classification problems~\cite{nimsuk2007improvement, nimsuk2008study} use features extracted by applying stepwise discriminant analysis (DA) to the short-time Fourier transform (STFT), followed by classification of these features with an LVQ (learning vector quantization) neural network. The Fourier power spectrum of time series $X(t)$ is defined by
$$
	\f_{w,\phi} X(t,\xi):=\left|\int X(\tau)w(t-\tau) \Exp{-i\xi\tau} d\tau \right|^2\ast\phi(t),
$$
where $w$ is a short-time low pass filter, $\phi(t)$ is a normalized smoothing window, and  $\xi$ is a frequency value. The Fourier power spectrum is defined as the expected value of square of the modulus of Short Time Fourier Transform(STFT $L_2$-moments)

\begin{equation}\label{fr:Fourier}
\bar{\f_{w}}(\xi) = \E_{t} \left[\f_{w,\phi} X(t,\xi)\right].
\end{equation}

Then, the STFT-based feature mapping of time series  $X(t)$ is
$$
\Phi: X \to \bar{\f_{\omega}}(\xi), \qquad \xi \in \Xi,
$$
where $\Xi$ is a given set of frequencies.

As the STFT feature extraction method constitutes prior art for our scattering approach, we applied it to our data as a performance benchmark.

Recent approaches~\cite{vergara2013performance,vembu2012time} use more elaborate statistical modeling of the data, as described below. Moreover, they have been applied to the dataset described in Section~\ref{ch:dataset} and are thus of greater interest to us. In this section we briefly describe these algorithms~\cite{vergara2013performance,vembu2012time}, which are used in Section~\ref{ch:Exper} as performance benchmarks.

The study~\cite{vembu2012time} models the time series at the output of the sensors, denoted by $X(t)$, using an auto-regressive linear model of order $p$ ~\cite{cuturi2011autoregressive,lutkepohl2007newTimeSeries}

\begin{equation*}
X(t) = C + \sum \limits_{i=1}^{p} A_i X(t-i) + e_t
\end{equation*}
where $e_{t}$ is a stochastic noise term, and $C,A_{1},\ldots,A_{p}$ are parameters determined from the given observations.

 The parameters represent the time series $C,A_{1},\ldots,A_{p}$  in the sense that the parameters can predict its value at time $t$ from past observations, allowing for an error $e_{t}$. Once these parameters have been estimated, we map each time series $X(t)$ to its feature vector by  $(C,A_{1},\ldots,A_{p})$ and use these vectors as the input for the classification step. The classification step in~\cite{vembu2012time} uses a kernel SVM with a Gaussian kernel function. The idea behind kernel SVMs is to use a function $k(x,x')$ that measures the similarity between the features corresponding to each pair of instances in the dataset. The most commonly used function is the Gaussian $k(x,x')=\exp(-\gamma\|x-x'\|^2),$  where  $\gamma$ hyper-parameter, which is usually learned by cross-validation. It can be shown that the kernel SVM algorithm is equivalent to mapping each feature vector to some high (possibly infinite) dimensional space, followed by linear partitioning of that space. Passing the points through a kernel function gives rise to non-linear decision boundaries, which cannot exist in linear classification. The results reported in~\cite{vembu2012time} pertain to a reduced dataset and simplified clarification/prediction, which used only four gases and source location only is classified to be either the right or left sides of wind tunnel.

The study~\cite{vergara2013performance} is of particular interest to us since the conditions were exactly the same as in our study, namely classification of $10$ gases with similar training scenarios. The features space is the maximum of the normalized response of the sensors of each board.
The classification scheme of~\cite{vergara2013performance} is based on an inhibitory SVM classifier with a Gaussian kernel, whose main concept is to train a classier $f_j$ for each possible label $j=1,\...,L$ where function  $f_j$  constitutes the distance between the correct label and the most offending incorrect answer. For a more detailed description, see~\cite{huerta2012inhibition}.

\section{Mathematical background}\label{ch:Math}

In this section we present the theoretical background for our approach. In Section ~\ref{ch:Wav_graph} we present a generalization of the critically sampled Haar wavelet transform on graphs~\cite{gavish2010multiscale}  to an over-complete representation. This transform allows us to exploit correlations between different sensing boards and plays a critical role in our feature extraction process. In Section~\ref{ch:scat}, we review the scattering convolution network, introduced by Mallat, which is one of building blocks of our method. Finally, in Section~\ref{sec:RF}, we provide some details on the Random Forest algorithm.

\subsection{Redundant wavelet decomposition on graph}\label{ch:Wav_graph}

Let $G=(V,E)$ be an undirected and unweighed graph, where $V=\{v_i\}_{i=1}^N$  is the vertex set and $E \subset V \x V$ is the edge set. The multiscale folder-decomposition ~\cite{gavish2010multiscale} $\vv=\{\v_j\}_{j=0}^{J}$ is collection of vertex set partitions where $\v^0=V$ is at resolution zero and for $j>0$, $\v^j=\{\upsilon^j_i\}_{i=1}^{n_j}, n_j=2^{-j}N$,  with

\begin{equation*}
\upsilon^j_i:=\upsilon^{j-1}_{\alpha_i} \cup \upsilon^{j-1}_{\beta_i}.
\end{equation*}
That is, the $i$-folder $\upsilon^j_i$ at level $j$ is obtained by grouping two folders $\upsilon^{j-1}_{\alpha_i}$  and $\upsilon^{j-1}_{\beta_i}$ from previous scale $j-1$.

%\subsection{Haar wavelet bases over the folder decomposition of a graph}
A function $X:G \to \mathbb{R}^{n}$ on a graph $G$ is defined by mapping each vertex $v \in V$ to $X(v) \in \mathbb{R}^{n}.$

The dot-product of two functions $X_1,X_2: G \to \mathbb{R}$, is defined by

\begin{equation}\label{fr:dotProdG}
\langle X_1,X_2 \rangle_{G} := \sum \limits_{v \in G} X_1(v) \cdot X_2(v).
\end{equation}

We now define an Haar wavelet ortho-basis transform over a folder decomposition $\vv$. The Haar wavelet is a function on the graph $G$ which is defined for each folder $\upsilon^j_i$ by
\begin{equation}\label{fr:HaarOrthoBasis}
\varphi_{j,i}:=\Ind_{v^{j-1}_{\alpha_i}} - \Ind_{v^{j-1}_{\beta_i}},
\end{equation}
where  $\Ind_{W}$ is the indicator function on the set $W\subset V$ of the graph $G$, given by

$$
\Ind_{W}(v)=
\left\{
  \begin{array}{l l}
    1 \text{ if } v \in W\\
    0 \text{ otherwise}
  \end{array} \right.
$$
According to~\cite{gavish2010multiscale}, the set of functions $\{\varphi_{j,i}\}$
is an orthogonal system defined on the graph $G=(V,E)$.
The set
\begin{equation}\label{fr:HaarBasis}
\Phi_{\vv}:=\{\varphi_{j,i}\cup \Ind_{v^j_{i}}\}_{i=0}^{n_j}  \text{, such that } n_j=2^{-j}N \text{ and }0\leq j \leq J
\end{equation}%_{i=0}^{2^{-j}d}}
defines  redundant wavelets on the graph $G.$
Note that on each level $j$, there are $2^{-j}N$  folders and that the maximum scale of a folder-decomposition is  $K_{\max}=\lfloor \log_2 N\rfloor$. The application of the Haar transform to a signal $X$ defined on a graph $G$ proceeds as follows. For first scale $k=$1, the Haar coefficients are
\begin{equation}
 \left\{
  \begin{array}{l l}
    X^1(i,0) = \langle \Ind_{v_i^1} , X \rangle_{G}, \\
    X^1(i,1)= \langle \varphi_{i,1}  , X \rangle_{G}.  \
  \end{array} \right.
 \end{equation}\label{fr:HaarZerroCoeef}

Then, for $k=2,...,K_{\max}$,
\begin{equation}\label{fr:HaarCoeef} \left\{
  \begin{array}{l l}
    X^k(i,2j) = \langle \Ind_{v_i^k}, X^{k-1}(i,j) \rangle_G, \\
    X^k(i,2j+1)= \langle \varphi_{i,k} , X^{k-1}(i,j) \rangle_G.
  \end{array} \right.
  \end{equation}
where $i=1\...2^{-k}d.$
Observe that when $X(t)$, $t \in \T$, is in fact a time series signal over the graph, then we use the notation $X^k(t,i,j)$ for the time dependent multiscale Haar coefficients.

\begin{figure}
  \centering
  \includegraphics[scale=0.5]{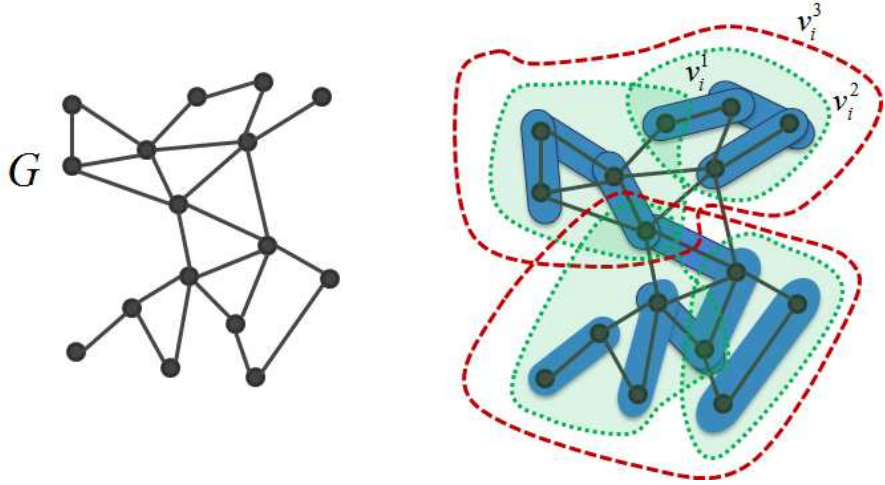}\\
  \caption{Multiscale folder-decomposition with overlapping.}\label{fg:RedundantHaar}
\end{figure}

Next, we introduce a generalization of the redundant wavelet transform~\cite{beylkin1992representation,shensa1992discrete} to functions defined on a graph. This approach is effective for signal processing and pattern recognition problems such as image denoising~\cite{ram2012redundant} and speech recognition~\cite{tohidypour2012new}. Our approach is to build a redundant scheme of wavelet decomposition by overlapping the folders at each level. Figure~\ref{fg:RedundantHaar} illustrates a multiscale decomposition with folder overlapping, which leads to an over-complete wavelet representation of a signal over a graph domain. It is easy to demonstrate that the system of redundant Haar-functions $\Phi_{\vv}$ defined in (\ref{fr:HaarBasis}), where ${\vv}$ is a folder decomposition with overlapping, is not necessarily an orthogonal basis. However, redundant Haar functions provide more mutual information between vertices.

\begin{figure}
  \centering
  \includegraphics[scale=0.125]{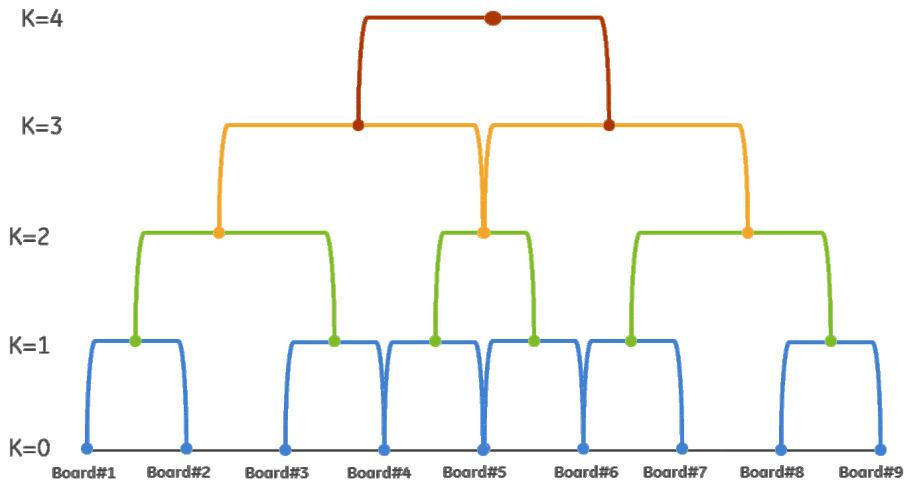}\\
  \caption{Multiscale folder-decomposition with overlapping Board Position}\label{fg:Lift_boards}
\end{figure}

For our olfaction problem, an overlapped folder decomposition using the neighborhood relationships of the boards. In Figure~\ref{fg:Lift_boards}, we can see the Haar wavelet lifting scheme, where overlapping is applied on center boards of the line position. Due to the wind direction in the tunnel, the air wind tunnel generates a diffusion chemical analyte. Therefore, centralized sensing boards contain more informative time series data~\cite{vergara2013performance}.

\subsection{Scattering convolution network on the graph time series}\label{ch:scat}
In this section we review scattering convolution networks and their adaptations for a graph time series.
In~\ref{subch:DeepLearn}, we review fundamentals of convolutional networks and deep learning. In \ref{subch:Scat}, we review the wavelet-based scattering network for time series. In \ref{subch:ScatStabl}, we emphasize the stability properties of the wavelet approach, which constitute an advantage over the Fourier power spectrum. In \ref{subch:ScatMoments}, we define feature extraction and classification methods based on scattering.

\subsubsection{Fundamentals of Convolution Networks and Deep Learning} \label{subch:DeepLearn}
Deep learning (DL) architectures~\cite{bengio2009learning}  are neural network-based algorithms modeled to mimic the functionality of the human nervous system. The DL methods have been successfully applied in a variety of pattern recognition problems such as computer vision (face recognition, image classification, and annotation), natural language processing, speech recognition, and audio representations signals.
%These DL algorithms perform successfully in a variety of pattern recognition problems.
Learning is achieved through hierarchical feature extraction of the observed data. The main idea is to apply nonlinear processing on the data that flows between the layers. In fact, hierarchical features are fitted weight parameters of neural networks (free parameters of each unit), which are calculated as the results of one complex optimization process.

The convolutional neural network (CNN)~\cite{lecun1998gradient} is one the most popular deep learning architectures for pattern recognitions problems for grid-based signals such as time series, images, and videos, such as ImageNet~\cite{krizhevsky2012imagenet}. The CNN consists of units that use overlapping patches of input signals to apply neurons~\cite{lecun2010convolutional}. In other words, the CNN learns a hierarchical net of convolutions (filters) of signals.

Mallat introduced a mathematical class of deep convolution networks \cite{mallat2012group}, which is called \emph{`Scattering Convolution Networks'}. The scattering convolution networks is unsupervised CNN architecture for grid-based signals that are obtained by cascading wavelet transforms and modulus pooling operators with the average of the amplitude of iterated wavelet coefficients. The scattering-based feature extraction is translation invariant and Lipschitz continuous to deformations~\cite{bruna2013invariant}.

When trying to apply DL/CNN techniques in problems such as ours, a difficulty arises since the geometric configuration of the sensors is not necessarily regular, as in image processing where the pixels are well aligned on a uniform grid. Thus, in this work, the concept of the uniform grid is replaced by a graph structure.

\subsubsection{Scattering convolution network} \label{subch:Scat}
In this work, the Scattering network is applied separately to each time series of the type $X(t)=X^k(t,i,j)$ computed by the graph analysis of Section \ref{ch:Wav_graph}. A scattering convolution network is obtained based on a cascade of wavelet convolution and modulus operators with smoothing operator (low-pass filter)  \cite{mallat2012group,bruna4104multiscale,bruna2013intermittent}. Let $\psi(t)$ be a complex wavelet, whose real and imaginary parts are orthogonal and have the same  $L_2$-norm, and $\int_{\R} \psi(t)dt=0$ with $|\psi(t)|=O\left( (1+t^2)^{-1} \right)$  with  dyadic dilations:
$$
\psi_j(t)=2^{-j}\psi(2^{-j}t), \quad \forall j \in \Z.
$$
The wavelet transform of time series $X(t)$  at scale $2^j$

$$
X \star \psi_j(t)=\int_{\R} X(u)\psi_j(t-u)du.
$$
We calculate the absolute value of the complex value coefficient
$$
U_1[j]X(t)=|X\star \psi_j(u)|.
$$
For each sequence of indices  $\bar{p}=(j_1,\...,j_m):j_1<j_2<\...j_m$ the \emph{order-$m$ scattering propagator} $U[p]$ is defined by:
$$
U_m[\bar{p}]X(t)=U[j_m]\...U[j_1]=||\...||X\star\psi_{j_1}|\star \psi_{j_2}|\...|\star \psi_{j_m}(t)|
$$

The \emph{windowed scattering} $S_{m,J}[\bar{p}]$ (scattering coefficients of order-$m$) is defined by
\begin{equation}\label{fr:ScatCoeff}
S_{m,J}[\bar{p}]X(t)=U[\bar{p}]X\star \phi_J=||\...||X\star\psi_{j_1}|\star \psi_{j_2}|\...|\star \psi_{j_m}(t)|\star \phi_J
\end{equation}
where $j_1<\...<j_m<J$ and $\phi_J(t)=2^{-J}\phi(2^{-J}t)$ is low-pass filter
with $2^J$ scale and $\int \phi(t)dt=1$.

The  scattering operator $S_J X(t)$ aggregates all scattering coefficients with order until layer $M$
\begin{equation}\label{fr:ScatAggregate}
S_J X(t)=\left( S_{m,J}X(t) \right)_{0\leq m\leq M}
\end{equation}
where $S_{0,J}X(t)= X \star \phi_J.$

\begin{figure}
  \centering
 \includegraphics[scale=0.125]{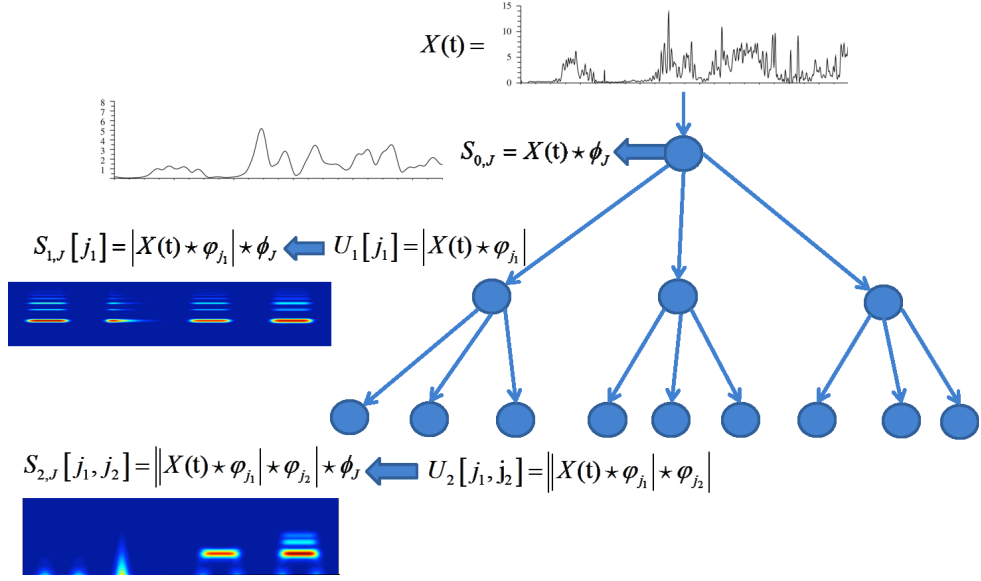}\\
  \caption{Two levels scattering convolution network of time series $X(t)$ with scattering propagators $U[j_1]$ and $U[j_1,j_2]$ } \label{fg:Scat_timeSer}
\end{figure}

The iterated procedure of scattering convolution network for a time series $X(t)$ is illustrated in Figure \ref{fg:Scat_timeSer}.

For most types of the signals, such as audio, images, biomedical signals, and finance time series, it is sufficient to compute the scattering coefficient of layers 0,1 and 2 (M=2).
$$
S_{2,J}X=
\begin{pmatrix}
S_{0,J}X\\
S_{1,J}\left[j_1\right]X\\
S_{2,J}\left[j_1,j_2\right]X
\end{pmatrix} _{j_1,j_2 \in \Lambda}
$$

\subsubsection{Scattering Deformation stability}\label{subch:ScatStabl}

The efficiency of a scattering representation comes from its invariance to local translations due to convolutions with $\phi_j$ and from its ability to linearize deformations, that is, its stability to time-warping.

The Fourier transform is unstable to deformation because dilating a sinusoidal wave yields a new sinusoidal wave of different frequency that is orthogonal to the original one~\cite{talmon2014manifold}. Let us define the deformation operator $D_{\tau}$  of signal $X(t)$

$$
D_{\tau}X(t)=X(t-\tau(t))
$$
where $\tau(t)$ non-constant deformation term.
As proven in~\cite[Theorem 2.12]{mallat2012group}, the scattering transform $S_J$ of a signal $X$ with compact support is Lipschitz continuous under action of deformation operator $D_{\tau}$.

\begin{equation}\label{fr:Scat_stable}
\|S_J(D_{\tau}X) -S_J(X)\|\leq C M\|X\| \left(2^{-J}|\tau|_{\infty} +|\nabla\tau|_{\infty}  \right)
\end{equation}
where  $|\tau|_{\infty}=\sup_t|\tau(t)|$ and $|\nabla\tau|_{\infty}=\sup_t|\nabla\tau(t)|<1.$

%\rvc{\begin{thm}\cite[Theorem 2.12]{mallat2012group} \label{Thn:Scat_stable}
%There exist $C>0$ such that for all $X(t)$ with $\|X(t)\|_1<\infty$ and all
%$\tau \in C^2$ with $\|\nabla\tau\|\leq 1/2$
%\begin{equation}\label{fr:Scat_stable}
%\|S_J(D_{\tau}X) -S_J(X)\|\leq Cm\|X\| \left(2^{-J}|\tau|_{\infty} +J(|\nabla\tau|_{\infty} + |H\tau|_{\infty} ) \right)
%\end{equation}
%where $|\tau|_{\infty}=\sup_t|\tau(t)|$ ,
%$|\nabla\tau|_{\infty}=\sup_t|\nabla\tau(t)|$ and $\|H\tau\|_{\infty}$ is Hessian sup-norm of $\tau$
%\end{thm}}{What do you expect the reader to gain from this theorem? Do you really need it? Can you do with just a simple explanation in words?}
This property guarantees stability of signals $X(t)$. It is clear that the deformation error is small if the scaling factor $J$ is $2^J\gg|\tau|_{\infty}$
and the signal $X(t)$ is smooth in $L_1$ meaning. In other words, the scattering metric satisfies invariance to local transformations and deformations.

\subsubsection{Scattering moments}\label{subch:ScatMoments}
As noted in Section \ref{subch:ScatStabl} above, the scattering network is stable under small deformation. It can therefore  be used as an effective feature space for many kinds of classification and regression problems.

State-of-the-art results of the scattering approach have been obtained for handwritten digit recognition and texture classification \cite{bruna2010classification} compared to  convolution neural networks (CNN)~\cite{lecun2010convolutional,ranzato2007unsupervised} and dictionary learning (DL)~\cite{mairal2012task}.

Scattering moments are defined as expected values over time of scattering coefficients, for each path  $\bar{p}=(j_1,\...,j_m):j_1<j_2<\...j_m$
$$
\bar{S}[\bar{p}]X = \E \left( S_{m,J}[\bar{p}]X \right) = \E \left(U_m[\bar{p}] X \right)
$$
For finite time series signal $|X(t)| = N$
$$
\bar{S}[\bar{p}]X = \frac{1}{N}\sum \limits_{t=1}^{N} ||X\star\psi_{j_1}|\star \psi_{j_2}|\...|\star \psi_{j_m}(t)|
$$
According to \cite{anden2013deep,bruna2010classification}, the standard way to build feature space based on scattering moments for finite time series signal $|X| = N$ is
\begin{equation} \label{fr:scat_moments}
\Phi: X \to \begin{pmatrix}
\bar{S}[j_1] X\\
\bar{S}[j_1,j_2] X
\end{pmatrix} _{j_1,j_2 \in \Lambda}
\end{equation}
where a scaling set
$ \Lambda = \{ (j_1,j_2) : 1\leq j_1 = 2^{z_1/Q_1}  \leq  N  \text{ and } 1 \leq j_2=2^{z_2/Q_2} \leq j_1 \: z_1, z_2 \in \Z \}$
defines a filter bank of scattering transform, such that  are number wavelets per octave of the first and the second layer.

Scattering moments have been used as features space\cite{anden2013deep} for time series classification problems: musical genre classification (GITZAN) and phone segment classification. In these kinds of signals, the best state-of-the-art results were obtained by an SVM classifier with Gaussian kernel.

\subsection{Random Forest}\label{sec:RF}
Random forest (RF)~\cite{breiman2001random} is a popular ensemble learning method for classification and regression. At training time, a diverse set of decision trees is constructed using randomization techniques. Their output is then averaged to overcome the potential bias of each tree. In some implementations, the decision trees are pruned to reduce variance. RF easily allows a  parallel architecture to be implemented in applications for testing and training scenarios. RF has been implemented in many recent applications for classification and regression problems~\cite{verikas2011mining}. In the current study, we found RF to be an effective classifier for our machine olfaction problem. Due to the fact that the RF approach is based on an ensemble of decision trees, the feature set consists of variables from different domains, including  categorical and continuous variables. This allows us to add a set of static parameters, such as airflow velocity and operating temperature, to the feature space of time series.

Let a set of vectors $X^{tr}=\{x_1\...x_n : x_i \in \R^d\}$
be a training set consisting of $n$ samples, each a $d$-dimension feature vector with associated response values $Y^{tr}=\{y_1,y_2,\...,y_n\}$ that can be categorical variables in a classification problem or continuous variables in a regression problem. The training data can be interpreted as samples governed by an unknown distribution from $X \times Y$.  The goal of the RF classifier is to learn (approximate) the oracle function $F:X \to Y$, such that learned function ̂$\hat{f}$ is the  best approximation of $F$ with respect to an error metric (loss function). In the tree bagging approach, for each $k=1\...N$, the algorithm randomly selects samples with replacement from the training set  $\{X^{tr}_k,Y^{tr}_k\}$ in order to build decision trees $\{T_k\}$. The final prediction function $\hat{f}$, is an average of the predictions of the decision tree $\{T_k\}$, or by weighted voting of the ensemble.

%\subsection{Tree Diversity}
%\emph{Feature bagging} is randomization of general scheme of decision  tree learning process that sample a random subset of the features  at each split of learned tree.
%
% Typically, for a data set with $d$ features, select $\sqrt{d}$ random features at each split of tree.
%
%The classical Breiman's random forest  algorithm combines tree bagging and feature bagging techniques using to it achieves higher \emph{'tree diversety'}, which help to minimize generalization error of obtained random tree ensemble \cite{kuncheva2003measures}. There exist many way to maximize diversity as follows: \emph{sample randomization} using special method of partitions of training set \cite{sinha2013efficient},which propose to use disjoint partitions of training dataset to
%train individual base decision trees, \emph{data perturbation} \cite{breiman2000randomizing} and  \emph{Feature randomization} e.g. Random Subspace \cite{ho1998random.}
% We propose to use feature randomization approach via  choosing of graph partition of the signal $\v \in \V_G$ and feature generation  $\bar{S}_{\v}x$ for each signal from training set $X_{tr}$ \ref{fr:Scat_TimeGraphFr}.

\section{Main algorithm}\label{ch:MainAlgo}
In this section, we present the main algorithm developed in this paper, in which all the elements introduced in the previous sections are brought together. The feature construction is performed using the {\bf Scattering Time Series on Graph (STSG)}.The computed feature vectors, along with the response variables, serve as the training set for an RF algorithm.
% Finally, in Section~\ref{subch:Scat_graph}  we define our version of Scattering Convolution Network for time series on the graph (STSG) that we use to compute the feature space of our machine olfaction algorithm.

%\subsection{STSG: Scattering time series on graphs}\label{subch:Scat_graph}
The STSG algorithm combines the Haar scattering transform on the graph and standard wavelet-based scattering net described previously.
The Haar scattering network architecture for the signal defined on graphs (not time series) was described in~\cite{chen2014unsupervisedHaarScatGraphs}. The scattering architecture is obtained by cascading multiscale Haar wavelet transform defined on an embedded subset (folders construction~\cite{gavish2010multiscale}), but without redundant wavelet decomposition.

$X$ denotes a multivariate time series defined on an unweighted graph domain $G=(V,E)$, with $\dim(V)=d$ and finite time domain $\T$

$$
X: G \x \T  \to \R.
$$

%\subsection{Mixed scattering transform of Time series sinal on graph domain}
The scattering time series on graphs $X(n,t)$ defined by:

\begin{equation}
	S_{J,\ell,k}[p_{\ell},\hat{j}]X=S_{J,\l}[p_{\ell}]X^k(t,i,\hat{j}).
\end{equation}\label{fr:Scat_TimeGraph}
$S_{J,\l}[p_{\ell}]$ consists of two actions. The first of these two actions is the calculation of $k$-levels at redundant Haar wavelet coefficients (\ref{fr:HaarCoeef}) with respect to the selected folder decomposition $\vv$. The architecture of folder decomposition $\vv$ can be derived from our understanding of the geometry of the sensing platform. The second action is the calculation of $\l$-levels scattering coefficients of time series such that $p_{\l}=j_1,\...j_l$, are their scaling paths (\ref{fr:ScatCoeff}).

The feature space of signal $X(n,t)$ consists of STSG moments, that is, the expected value scattering time series on graphs transform $S_{J,\ell,k}[p_{\ell},\hat{j}]$ over time.

 \begin{equation}\label{fr:Scat_TimeGraphFr}
 \Phi: X \to \bar{S}_{\v} =
  \mathbb{E}_{t} \left[S_{J,\ell,k}[p_{\ell},\hat{j}]X(t,.)\right].
 \end{equation}

%\subsection{Classification algorithm}\label{subch:STSGAlgo}
Our machine learning task is as follows: Given a set of training signals $X^{tr},Y^{tr}$, such that each instance is a time series defined on the same graph $G$, the learning algorithm must seek a prediction function $f: X \to Y$.

Equipped with our training set, we calculate STSG moments  $\bar{S}_{\v}(x_i)$, $x_i \in X^{tr}$ (see \ref{fr:Scat_TimeGraphFr}). The second step is applying dimensionality reduction of the STSG moments to $d$ principal components
$$
\Phi: x_i \to \Phi^{d}(\bar{S}_{\v}(x_i)), \forall x_i \in X^{tr}.
$$
The dimensionality reduction increases stability. The scattering domain lies on a low dimensional manifold~\cite{bruna2010classification}.

In some of the classification scenarios, we have additional information about each instance $x_i$, which it is added to the feature space.  Since our learning algorithm is based on an RF classifier, with trees constructed over bagging of the training set, each iteration $k$ begins with random sampling $\{X_k,Y_k\}$ from the full training set, followed by feature mapping, $\bar{S}_{\v}(x_i)$, $x_i \in X_k$.

\begin{algorithm}
\caption{Training Random Forest of Scattering Graph Net}\label{algo:clasif}
\begin{algorithmic}[1]
\Procedure{TrainRF}{$X^{tr},Y^{tr}$}

\For{$k\gets 1, T$} \Comment{Loop for ensemble trees}

\State{$(X_k,Y_k) \gets \textbf{Bagging}(X^{tr},Y^{tr})$} \LONGCOMMENT{\textbf{Bagging}: Random sampling with replacement}

%\State $\v_t  \gets \textbf{RandHaarGraphPartion}(X_t)$ \LONGCOMMENT{Random chose a one of possible folders partition $\V_G$}

\State $\bar{S}_{\v}(x_i) \gets \textbf{STSG}(x_i), x_i \in X_k$ \LONGCOMMENT{Computing STSG moments}

\State $(\Phi^d_k,P_k^d) \gets PC_d\{\bar{S}_{\v}(x_i), x_i \in X_k \})$ \LONGCOMMENT{\textbf{Features space}: Dimensionality reduction to $d$ principal components}

\State $f_k \gets \textbf{TreeGrow}(\Phi^d_k,Y_k)$ \LONGCOMMENT{Growing a decision tree with feature bagging}

\State $F \gets F +f_k$ \LONGCOMMENT{Ensemble tree building}
\EndFor
\State $F \gets \frac{1}{N}F$\LONGCOMMENT{Final classifier uniform normalization}

\State \textbf{return} $F$\LONGCOMMENT{Output: Final tree ensemble and learned PCA transform}
\EndProcedure
\end{algorithmic}
\end{algorithm}

\section{Results} \label{ch:Exper}
In this section we demonstrate the application of our approach to machine olfaction problems.
In Section \ref{subch:features} , we  review the raw feature space. In Section \ref{subch:scenarios} we define classification scenarios.
 Finally, in Sections \ref{ch:Clasif}, \ref{ch:CO_consentr} and \ref{ch:SourceLoc}, we compare our results with prior-art scattering techniques and state-of-the-art machine olfaction techniques described in Section ~\ref{sec:sateOFart}.

\subsection{Feature Space}\label{subch:features}

To the STSG features (see Section~\ref{ch:MainAlgo}), we add some of the pre-established conditioning parameters described in Section~\ref{ch:dataset}

\begin{itemize}
  \item \emph{Heater voltage} $V_H \in \{4.0V, 4.5V, 5.0 V, 5.5 V, 6 V \}$,
  \item \emph{Airflow velocity} measured in rotation per minute: $rmp \in \{1500, 3900, 550\}$,
  \item \emph{Nominal  concentration} measured  in  parts-per-million  by  volume(ppmv).
\end{itemize}

The features of ambient temperature and relative humidity are not included since they have very low variance.  Our experiments clearly show that adding these static features significantly improves the performance of the prediction algorithm.

In addition, the location of each time series is known, namely, the position and board number with respect to the chemical source. The location of the sensing board is

\begin{equation} \label{fr:loc}
F_{loc}=\{X_{pos} \times X_{bord}\},
\end{equation}
where
\begin{itemize}
  \item $X_{pos} \in \{0.25, 0.5, 0.98, 1.18, 1.40, 1.45 \}$,
  \item $X_{board}= 0.13:0.13:1.2$.
\end{itemize}

\subsection{Classification scenarios}\label{subch:scenarios}
Our results are for complex scenarios of gas classification and source detection. These scenarios are motivated by proposals of the Department of Defense (DoD) for the development of chemo-sensing solutions and standards for early warning and protection of military forces against potential chemical and biological attacks (see \cite{guideHazards},\cite{chemHazards}). Specifically, the scenarios are as follows:

\begin{itemize}
  \item Gas classification problem (10 labels),
  \item Gas concentration prediction for $CO$ only (binary classification problem: $1,000$ ppm and $4,00$ ppm only),
  \item Source localization problem – prediction of source location with respect to local coordinates in the wind tunnel.
\end{itemize}

In all the above scenarios, we applied learning with/without static features and with/without source location information.

Based on prior studies, we constructed two ways to aggregate raw features:
\begin{enumerate}
  \item \textbf{Board Column}\cite{vergara2013performance}  aggregating all nine boards from than same position to a $72$-dimension time series,
  \item \textbf{Single Board}\cite{vembu2012time} using only the $8$-dimension time series from a single board.
\end{enumerate}
The `Single Board' scenario is significantly more difficult because fewer features are used.
We compared our results with frequency-domain features based on short-time Fourier transformation (\ref{fr:Fourier}) and simple features based first two statistical moments of each time series of sensor response.\\
In other words, we  compared three groups of features:
\begin{enumerate}
  \item STSG features (\ref{fr:Scat_TimeGraphFr}),
  \item Statistical moments,
  %\item AR features \ref{fr:AR_features}
  \item Fourier Power spectrum(STFT $L_2$-moments) (\ref{fr:Fourier}).
\end{enumerate}
For these three groups of features, we applied the same RF classifier with $200$ grown trees. The error rate was calculated as expected value and variance of 5-folder cross validation.

\subsection{Gas classification results}\label{ch:Clasif}

 Table \ref{tbl:Classif_byPos} shows the classification performance of Board Column suitable for the conditions  \cite{vergara2013performance}. The performance of the random forest classifier for all three groups of features is significantly better than the mean results of a previous study \cite{vergara2013performance}. As shown, using our STSG features provides significantely better performance in the more complex scenarios.

\begin{table}[h]
	\centering
\caption{\textbf{Classification performance of \emph{`Board Column'} scenario}.Testing error rate RF models with proposed STSG-algorithm, Short Time Fourier Transform(STFT) and statistical moments features, with/withput static features and with/withput location information  } % title of Table
\begin{tabular}{c|c||c|c|c|c}
\hline\hline
\begin{tabular}[c]{@{}c@{}}Static\\ feature\end{tabular} & Location & STFT           & \begin{tabular}[c]{@{}c@{}}Statistical\\ moments\end{tabular} & STSG                 & \begin{tabular}[c]{@{}c@{}}\cite{vergara2013performance}\\ \end{tabular}\\[0.5ex]  \hline\hline
\multirow{2}{*}{FALSE}                                   & FALSE    & 2.52\%($\pm$0.09\%) & 0.574\%($\pm$0.07\%)                                             & {\bf 0.47\%($\pm$0.07\%)} & \multirow{4}{*}{7.89\%}                                           \\ %\cline{2-5}
                                                         & TRUE     & 2.41\%($\pm$0.16\%) & 0.58\%($\pm$0.06\%)                                              & {\bf 0.48\%($\pm$0.03\%)} &                                                                   \\ %\cline{1-5}
\multirow{2}{*}{TRUE}                                    & FALSE    & 1.26\%($\pm$0.15\%) & {\bf 0.09\%($\pm$0.07\%)}                                        & 0.26\%($\pm$0.08\%)       &                                                                   \\ %\cline{2-5}
                                                         & TRUE     & 1.31\%($\pm$0.25\%) & {\bf 0.09\%($\pm$0.02\%)}                                        & 0.24\%($\pm$0.07\%)       &                                                                   \\ \hline\hline
\end{tabular}
\label{tbl:Classif_byPos}
\end{table}

A comparison in the \emph{'Single Board'} scenario is presented in Table \ref{tbl:Classif}.

\begin{figure}
  \centering
  \includegraphics[scale=0.5]{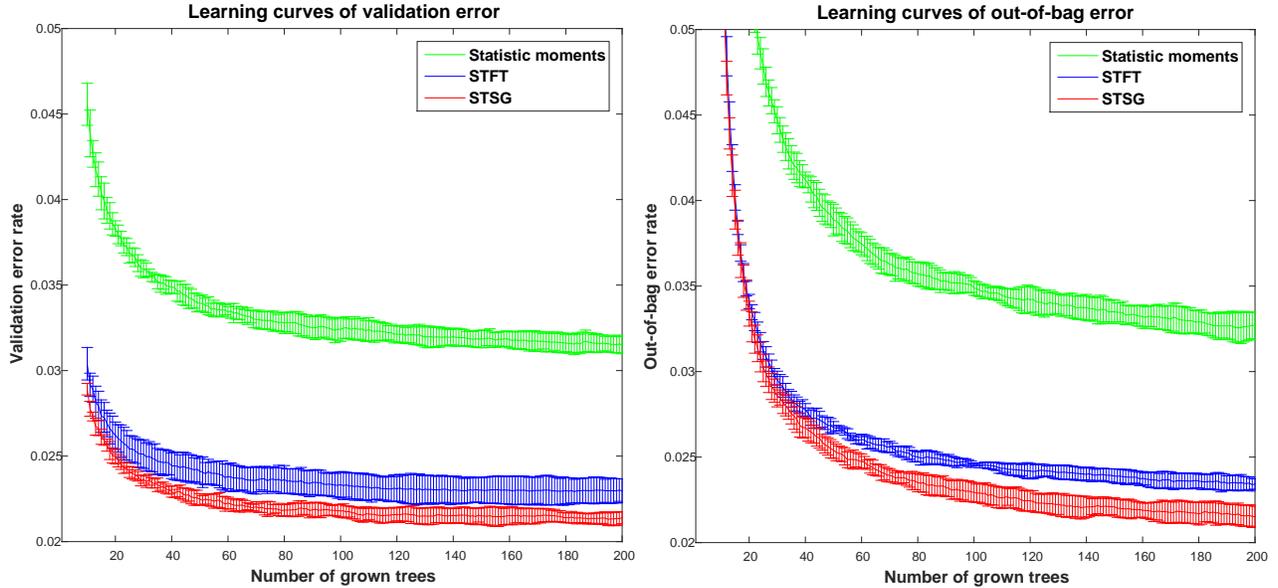}\\
  \caption{Learning curves of the validation error and out-of-bag errors of classification performance of Single Board scenario scenario.}\label{fg:Learn_rate_classif}
\end{figure}

\begin{table}[h]
\centering
\caption{\textbf{Classification performance of \emph{`Single Board'} scenario}. Testing error rate RF models with STSG, STFT and statistic moments features with/without static and location information } % title of Table
\begin{tabular}{c|c||c|c|c}
\hline\hline
\begin{tabular}[c]{@{}c@{}}Static \\ features\end{tabular} & Location & STFT            & \begin{tabular}[c]{@{}c@{}}Statistic\\  moments\end{tabular} & STSG            \\[1ex]  \hline\hline
\multirow{2}{*}{FALSE}                                     & FALSE    & 19.68\%($\pm$0.37\%) & 24.96\%($\pm$0.15\%)                                              & \textbf{18.11\%($\pm$0.11\%)} \\[0.5ex] %\cline{2-5}
                                                           & TRUE     & 14.74\%($\pm$0.20\%) & 18.20\%($\pm$0.11\%)                                              & \textbf{13.63\%($\pm$00.20\%)} \\[0.5ex] %\hline
\multirow{2}{*}{TRUE}                                      & FALSE    & 3.15\%($\pm$0.12\%)  & 4.78\%($\pm$0.08\%)                                               & \textbf{2.80\%($\pm$00.08\%)}  \\[0.5ex] %\cline{2-5}
                                                           & TRUE     & 2.30\%($\pm$0.07\%)  & 3.15\%($\pm$0.05\%)                                               & \textbf{2.14\%($\pm$0.03\%) } \\[0.5ex] \hline\hline
\end{tabular}
\label{tbl:Classif}
\end{table}
Figure \ref{fg:Learn_rate_classif} shows error bars on learning curves of validation error and out-of-bag error over the number of grown classification trees in the random forest ensemble for each kind of features.

 The STSG learning curves have the fastest decay rate. This proves the learnability of the proposed method.

\subsection{Detection of $CO$-concentration }\label{ch:CO_consentr}
We now address a binary classification problem, in which the goal is to determine concentration of the $CO$ substance (carbon monoxide). Note that the given dataset includes every chemical substance in the experiment in only one nominal concentration, except for the carbon monoxide, which was collected in two different concentrations of $1,000$ ppm and $4,000$ ppm. We built a subset with binary labeling from a given dataset that contained only experiments with $CO.$ We then applied our RF classifier for compared feature space.
Table~\ref{tbl:ClassifCO} presents the classification performance of the Single Board scenario of the binary classification problem. Note that,  in this scenario, we cannot compare results with static features because these features have been simulated with different unique static parameters.

\begin{table}[h]
\centering
\caption{\textbf{Classification performance of  \emph{'Single Board} scenario for $CO$ concentration}. Testing error rate of RF classifier of proposed STSG, STFT, and statistical moments features space with/without location information .} % title of Table
\label{tbl:ClassifCO}
\begin{tabular}{c||c|c|c}
\hline\hline
Location & STFT & \begin{tabular}[c]{@{}c@{}}Statistic \\ moments\end{tabular} & STSG \\ \hline\hline
FALSE & 0.30\%($\pm$0.05\%) & 1.00\%($\pm$0.16\%) & \textbf{0.24\%($\pm$0.07\%)} \\% \hline
TRUE & 0.26\%($\pm$0.06\%) & 0.61\%($\pm$0.10\%) & \textbf{0.20\%($\pm$0.05\%)} \\ \hline\hline
\end{tabular}
\end{table}
The proposed STSG feature space clearly demonstrates performance that is superior to that of other methods.. We did not apply this problem ’Board Column’, because ’Board Column’ scenario for all gases has perfect performance yet.

\subsection{Source localization }\label{ch:SourceLoc}
The source localization scenario is a learning of regression model for high accuracy detection of MOX-sensors location with respect to the gas substance. Tables \ref{tbl:LocPred_pos} and \ref{tbl:LocPred} show the rate error of the compared features space. The two types of features aggregation, ’Board Column’ and ’Single Board’ scenarios, were used respectively.
The error rate is the  $L_2$-norm error in meters between sensor board location and predicted location.
Note that if we use feature aggregation by Single Board, the classifier predicts the 2D location of the sensing board $F_{loc}$ (\ref{fr:loc}). When feature aggregation by Board Column is used, the classifier predicts distance of line position only $X_{pos}$.

\begin{table}[h]
\centering
\caption{\textbf{Location prediction performance of \emph{'Board Column'}}: $L_2$-norm error in meters with respect to source location}
\label{tbl:LocPred_pos}
\begin{tabular}{c||c|c|c}
\hline\hline
\begin{tabular}[c]{@{}c@{}}Static\\ Features\end{tabular} & STFT & \begin{tabular}[c]{@{}c@{}}Statistic\\ moments\end{tabular} & STSG \\ \hline\hline
FALSE & 0.02155($\pm$0.00063) & 0.00745($\pm$0.00028) & \textbf{0.00477($\pm$0.00060)} \\\hline
TRUE & 0.02025($\pm$0.00076) & 0.00773($\pm$0.00032) & \textbf{0.00481($\pm$0.00043)}\\ \hline\hline %inserts single line
\end{tabular}
\end{table}

\begin{table}[h]
		\centering
\caption{\textbf{Location prediction performance of \emph{'Single Board} scenarios}: $L_2$-Norm error in meters with respect to source location} % title of Table
\begin{tabular}{c||c|c|c}
\hline\hline
\multicolumn{1}{c||}{\begin{tabular}[c]{@{}c@{}}Static \\ features\end{tabular}} & \multicolumn{1}{c}{STFT} & \multicolumn{1}{|c|}{\begin{tabular}[c]{@{}c@{}}Statistic \\ moments\end{tabular}} & \multicolumn{1}{c}{STSG} \\[1ex]  \hline\hline
FALSE                                                                          & 0.174($\pm$0.001)             & 0.288($\pm$0.001)                                                                     & {\bf 0.162($\pm$0.001)}       \\ \hline
TRUE                                                                           & 0.161($\pm$0.010)             & 0.280($\pm$0.006)                                                                     & {\bf 0.150($\pm$0.008)}
\\ \hline\hline %inserts single line
\end{tabular}
\label{tbl:LocPred}
\end{table}

\section{Conclusions}
In the current study we presented a novel methodology that can be used for pattern recognition problems in which the signals are collected from an array of possibly non-uniform ensemble of sensors. We defined a novel scattering transform for multidimensional time series on the graph (STSG), which used redundant wavelet graph decomposition. In this study, we focused on machine olfaction problems in which the dataset was obtained from a chemical gas sensor array in a turbulent wind tunnel. We applied our methodology to three machine learning problems: classification of $10$ different gases at various concentrations, detection of $CO$ consecration, and chemical substance localization.\\
% Also we proposed novel transfer learning approach adapted for real life scenario of machine  olfaction  applications.
The next step of this research is to develop the proposed method formachine olfaction in cases involving  turbulent gas mixtures~\cite{fonollosa2014chemical}.
Another interesting research direction would be to add \textit{soft supervised learning to scattering net}. Recent advances in classical deep learning, such as CNN, have demonstrated high accuracy performance. These methods are based on the learning of convolutional filters. However, it must be noted that deep learning architecture typically requires a large training dataset and the learning process is computationally intensive. Recall that the learning process when using scattering networks is faster since we use pre-designed filters. One could then consider `soft' learning variant of the scattering network, where standard wavelet filters are used as initial filters and then optimization is applied to only a few wavelet parameters.

%%%%%%%%%%%%%%%%%%%%%%%%%%%%%%%%%%%%%%%%%%%%%%%%%%%%%%%%%%%% 